\documentclass[runningheads,a4paper]{svproc}
\usepackage[latin1]{inputenc}
\usepackage{amssymb}
\usepackage{graphicx}
\usepackage{amsmath} 
\usepackage{subfigure}

\usepackage{verbatim}
\usepackage{floatrow}
\usepackage{multicol,caption}
\usepackage{multirow}
\newfloatcommand{capbtabbox}{table}[][\FBwidth]

\DeclareCaptionLabelFormat{andtable}{#1~#2  \&  \tablename~\thetable}
\usepackage{lipsum}
\usepackage{etoolbox}

\usepackage{url}
\urldef{\mailsb}\path|{ogil, sanfeliu}@iri.upc.edu|
\newcommand{\keywords}[1]{\par\addvspace\baselineskip
\noindent\keywordname\enspace\ignorespaces#1}

\begin{document}

\mainmatter  

\title{Robot Navigation Anticipative Strategies in Deep Reinforcement Motion Planning }

\titlerunning{Robot Navigation Anticipative Strategies in DRL Motion Planning}

\author{\'{O}scar Gil\thanks{Work supported under the Spanish State Research Agency through the Maria de Maeztu Seal of Excellence to IRI (MDM-2016-0656) and ROCOTRANSP project (PID2019-106702RB-C21 / AEI / 10.13039/501100011033). \'{O}scar Gil is also supported by the Spanish Ministry of Science and Innovation under an FPI-grant, BES-2017-082126.}, and Alberto Sanfeliu}

\institute{Institut de Rob\`otica i Inform\`atica Industrial, CSIC-UPC\\
\mailsb}
\authorrunning{}

\maketitle

\vspace{-5mm}

\begin{abstract} 

The navigation of robots in dynamic urban environments, requires elaborated anticipative strategies for the robot to avoid collisions with dynamic objects, like bicycles or pedestrians, and to be human aware. We have developed and analyzed three anticipative strategies in motion planning taking into account the future motion of the mobile objects that can move up to 18 km/h. First, we have used our hybrid policy resulting from a Deep Deterministic Policy Gradient (DDPG) training and the Social Force Model (SFM), and we have tested it in simulation in four complex map scenarios with many pedestrians. Second, we have used these anticipative strategies in real-life experiments using the hybrid motion planning method and the ROS Navigation Stack with Dynamic Windows Approach (NS-DWA). The results in simulations and real-life experiments show very good results in open environments and also in mixed scenarios with narrow spaces.

\keywords{Deep Reinforcement Learning, Robot Navigation, Social Force Model, Human-Robot Interaction, Anticipation}
\end{abstract}

\vspace{-7mm}
\section{Introduction}\label{sec_introduction}
Collaborative tasks \cite{ajoudani2018progress} are essential in areas like social robotics and human-robot interaction. Usually, social-aware robot navigation \cite{charalampous2017recent} is involved in these tasks in different ways like, for example, assistive robots \cite{urdiales2011new}, approaching task, collaborative searching \cite{marc_searching} or side-by-side navigation \cite{repiso2020people}.

The most typical robot navigation approaches take into account the map knowledge to compute a global plan to a goal avoiding obstacles in combination with a local planner to react in a short range. Dynamic Windows Approach (DWA) \cite{fox1997dynamic} or Social Force Model (SFM) \cite{helbing1995} are reactive local planners and they are useful in presence of unpredictable situations in dynamic environments. However, when there are moving obstacles (e.g., person, bicycles, scooters, etc.), reactive behaviors are not enough. It is more useful to have an anticipative behavior to deal with them \cite{ferrer2019anticipative}.

Due to the noise and the incomplete environment knowledge, planners which require accurate data like Artificial Potential Fields (APF) \cite{chiang2015path} can fail in real life applications. In these cases, Deep Reinforcement Learning (DRL) can provide a feasible solution through algorithms like Deep Deterministic Policy Gradients (DDPG) \cite{lillicrap2015continuous} and Proximal Policy Optimization (PPO) \cite{han_yiheng}. These approaches yield good performance in robot navigation without complete map knowledge. 

\begin{figure}[t]
    \centering
    \includegraphics[width=0.5\linewidth]{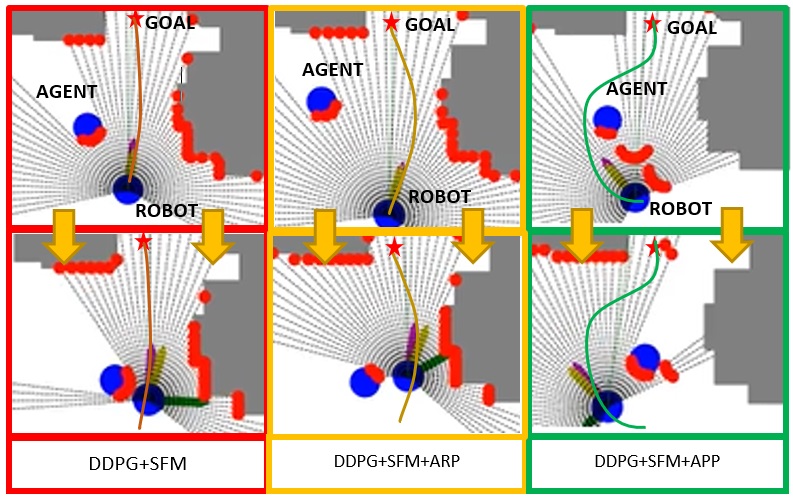}
    \caption{{\bf Comparison between models.} The image shows the trajectory changes when two anticipative strategies are applied in the same situation. The ARP model follows a more secure trajectory than the hybrid model and the APP model chooses a trajectory to avoid collisions passing from behind the agent.}
\label{example_model}
\end{figure}

In this work, we present three different anticipative strategies that can be used in different motion planners. We have tested these strategies using our hybrid motion planning model for robot navigation that combines a DDPG policy based on \cite{chiang2019learning} and SFM. The three models denominated "Anticipative Turn (AT)", "Anticipative Robot and Pedestrian's Propagation (ARP)" and "Anticipative Pedestrian's Propagation (APP)", compute the best hybrid navigation decision taking into account the prediction of the mobile objects in open and complex dynamic scenarios (refer to Fig. \ref{example_model}). We have validated the models through real-life experiments and using simulation in dynamic scenarios for different noise values and moving obstacle velocities. 

The remainder of this paper is organized as follows. In section 2, the related work is introduced with a description of the hybrid model that combines DDPG and SFM. The anticipative strategies are described in section 3. In Section 4, a quantitative and qualitative evaluation is presented through simulations and real-life experiments. Finally, in section 5, the conclusions are provided.

\section{Related Work}

\subsection{AutoRL for Robot Navigation}
Automated Reinforcement Learning (AutoRL) \cite{chiang2019learning,faust2019evolving} is a training algorithm that uses DDPG or other DRL approaches \cite{mnih2015human,haarnoja2018soft} to train policies searching for the reward and neural network optimal hyperparameters that maximize an objective function.

In \cite{chiang2019learning}, the algorithm is applied to point-to-point navigation task (P2P), which consists on navigating from an initial position to a goal using as policy actions, the velocities at each navigation step. The environment information is the robot Lidar observations in goal polar coordinates. The reward function is:
\begin{equation}
    R_{\theta_{r_{\mathrm{P2P}}}}=\boldsymbol{\theta}_{r_{P2P}}^{T}[r_{\mathrm{step}} \, r_{\mathrm{distg}} \,  r_{\mathrm{col}} \,  r_{\mathrm{turn}} \,  r_{\mathrm{clear}} \,  r_{\mathrm{g}}]
\end{equation}
where $\boldsymbol{\theta}^{T}_{r_{\mathrm{P2P}}}$ are the reward hyperparameters for the P2P task, $r_{step}$ is a constant penalty at each step in the episode with value 1, $r_{distgoal}$ is the negative Euclidean distance to the goal, $r_{col}$ is 1 when there are collisions and zero otherwise. $r_{turn}$ is the negative angular speed, $r_{clear}$ is the distance to the closest obstacle and $r_{g}$ is 1 when the agent reaches the goal and zero, in the remaining cases.

AutoRL have been implemented as a local planner that complements a global planner as PRM \cite{francis2020long} in large indoor environments. The results in real environments show high performance and robustness to noise. However, AutorRL requires a very large training time and it has high computational cost.

\subsection{Hybrid Policy with SFM}
This subsection explains the hybrid model proposed in \cite{oscar_anais_sensors2021} resulting from a DDPG training and the SFM with collision prediction (CP) \cite{zanlungo2011}. 

First, the model is trained using the DDPG algorithm with the same noise and optimal hyperparameters, obtained in \cite{francis2020long} for a differential drive robot model. The training is performed with only static obstacles.

This algorithm uses feed-forward networks for both the actor and the critic. The actor network is the policy, $\pi({\bf o};\theta_{a})$, with network parameters $\theta_{a}$, that takes the environment observations ${\bf o}$ and provides an action (linear and angular velocities): $\pi({\bf o};\theta_{a})=(a_{l},a_{\phi})$.

Second, in the evaluation phase, the people velocities obtained using a tracker \cite{vaquero2019robust,linder2016multi} are used in the SFM with CP to compute velocity changes $({\bf \Delta v})_{rep}=(\Delta v_{l}, \Delta v_{\phi})$ as is explained in \cite{oscar_anais_sensors2021}. These velocity changes are added to the DDPG actions to calculate the actions provided by the hybrid model:

\begin{equation}
    (a'_{l}, a'_{\phi})  = (a_{l}, a_{\phi}) + (\Delta v_{l}, \Delta v_{\phi})
\end{equation}

This hybrid model takes the agent velocities as additional entries regarding the pure DDPG model. Another option is to exploit this information as an entry for training a new DDPG model. This option has several disadvantages like searching for the new optimal hyperparameters for the implementation and a longer time to learn the policies in a more complex environment.

\subsection{Anticipative Planning}

Anticipation is the ability to react taking into account not only the actual situation, but also a prediction in a future time window. It can be useful in human-aware motion planning to avoid dangerous situations in the future, where finding a solution could be very difficult or maybe impossible.

Some works \cite{cosgun2016anticipatory,ferrer2019anticipative} have implemented anticipation in path planning based on human trajectory prediction \cite{rudenko2020human}. There are several methods to predict trajectories based on previous human positions or other environment cues. Despite the unpredictable and multimodal nature of the problem, the constant velocity model (CVM) \cite{scholler2020constant} gives a good approximation in many cases.

\section{Anticipative Models}
The hybrid model described in the previous section avoids collisions but when the robot is close to the person. This section explains the three anticipative strategies proposed in this paper to avoid collisions in a human-aware way.

\subsection{Problem Formulation}
As in \cite{chiang2019learning}, this work tackles the robot P2P navigation task as a Partially Observable Markov Decision Process (POMDP). The action space, $A$, includes the linear and angular velocities for a unicycle robot model: ${\bf a}=(a_{l},a_{\phi})$ where $a_{l} \in [-0.2,1]$ m/s and $a_{\phi} \in [-1,1]$ rad/s. The robot and the agents or moving obstacles (e.g., pedestrian) are modelled in the environment as circles of radius $R_{r}$ and $R_{p}$ respectively. The reward function is the one described in (1).

The observations, ${\bf o}=({\bf o}_{l},{\bf o}_{g})^{\theta_{n}} \in O$, are divided in two parts: ${\bf o}_{l}$, includes the 64 1-D Lidar distance measures between 0 and 5 m, which are taken during the last $\theta_{n}$ simulation steps and ${\bf o}_{g}$ includes the goal polar coordinates. The Lidar field of view is $220^{o}$.

\subsection{Anticipative Turn (AT)}
This approach propagates the robot during a time-step using the actual velocity command before moving the robot with that command, and checks the possible collisions in the next movement with the last Lidar's points. 

In case of detection of a future collision, the robot does not execute the velocity command given by the hybrid model, but changes it for a command sequence of 6 steps. During the first 4 steps, the robot only turns using the angular velocity in a random way. In the next 2 steps, the robot moves only with the linear velocity. If a collision is detected when the robot movement is propagated the last 2 steps, the sequence is initialized until the robot can execute the complete sequence without collision.

\subsection{Anticipative Robot and Pedestrian's Propagation (ARP)}

The AT approach has been designed for static environments. For this reason, it only works well when the velocity of the moving agents is very low or with only static obstacles. For these reasons, the AT model is not activated when a possible collision with moving obstacles is detected.

To avoid static and moving obstacles the ARP model is proposed, where at each step, the robot and the moving agents are propagated during $n$ time-steps in a local map. In this propagation, the static obstacles and moving agents of the environment are used to build the local map using the information of the last Lidar's observations. The robot motion is propagated using the hybrid policy and the moving agents are propagated using the CVM. The propagation ends before the $n$ steps when there is a collision or the goal is reached. 

After the propagation, the action ${\bf a}(t)=(a_{l}(t),a_{\phi}(t))$ that will be taken by the robot at the present time $t$ is the same as the action taken when the robot was at the minimum distance of the obstacle during the propagation phase. This action increases the probability of avoiding a collision $n$ steps before it occurs.

\subsection{Anticipative Pedestrian's Propagation (APP)}

The ARP model, described in the previous section, requires that the local map, with the static and dynamic objects, has to be modified taken into account the Lidar's future propagation measurements. In some cases, the future Lidar's measurements cannot detect objects, because of partial occlusions in the present time, before the robot or agent's propagation. For these reasons, the model in this subsection does not create a local map or propagate all the environment.

The APP model considers a constant velocity prediction for all moving agents detected by the Lidar in the next $n$ time-steps. Each new predicted agent position has an associated uncertainty that can be modelled as a two-dimensional Gaussian distribution $N(${\bf c}$,\Sigma)$, centered in the position ${\bf c}$ with variance $\Sigma$. 

A circle centered in ${\bf c}$ with radius, $r=\Sigma$, represents the area with more probability to localize the agent. If we consider that the uncertainty of the agent's motions at each position is independent from other agent's motions, the propagation of the uncertainty and the radius during the $n$ steps is the sum of the variances, $\Sigma_{n}=n\Sigma$.

The total radius $r_{n}$ of the anticipative circle after $n$ steps is the agent radius $R_{p}$ plus the uncertainty $\Sigma_{n}$:
\begin{equation}
    r_{n}=\Sigma_{n} + R_{p}
\end{equation}
The APP model considers anticipative circles as virtual obstacles detected using the Lidar. At each step, the robot detects all the anticipative circles of each one of the agents for the next $n$ time-steps. If the robot collides with a circle, it is not considered a real collision. That circle and the next circles disappear and only the circles from the present time to the circle previous to the collision remain.

\begin{figure*}[t]
    \centering
    \includegraphics[width=0.97\linewidth]{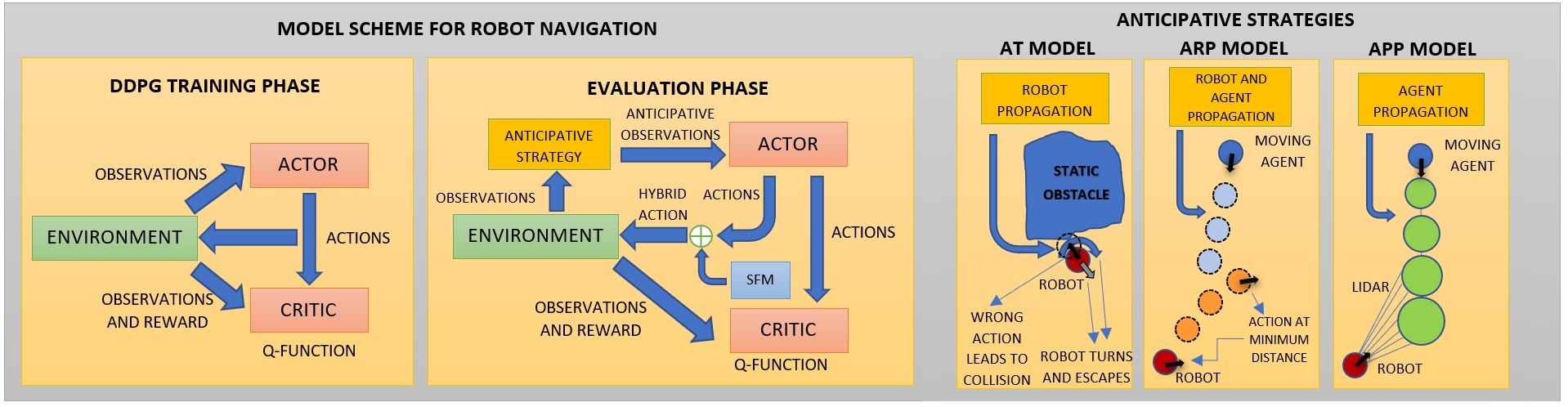}
    \caption{{\bf Model Scheme.} Training is performed without SFM. In evaluation, SFM actions and anticipation can be added.}
\vspace{0mm}
\label{model}
\end{figure*}

A scheme of the approaches is shown in Fig. \ref{model}.

\section{Simulations and Experiments}

\begin{figure*}[t]
    \centering
    \subfigure[Training Map 24x20 m]{\includegraphics[height=0.25\textwidth]{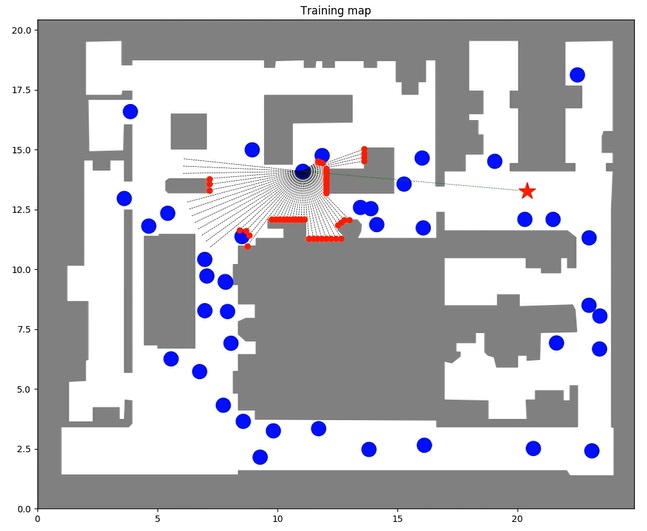}}
    \subfigure[Building 2 60x47 m]{\includegraphics[height=0.25\textwidth]{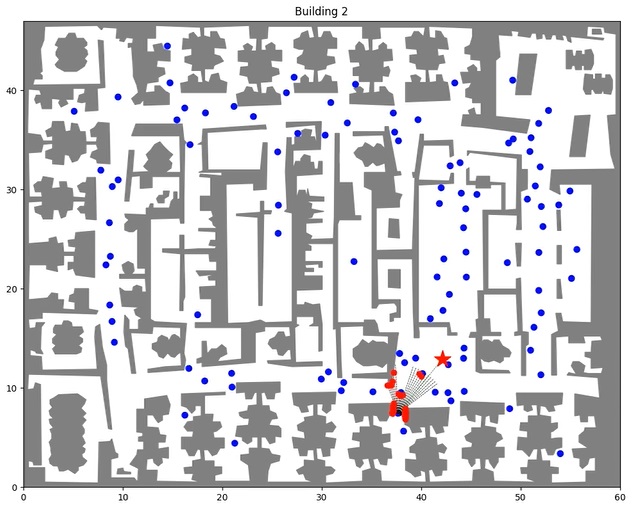}}
    \subfigure[Empty Map 20x20 m]{\includegraphics[height=0.28\textwidth]{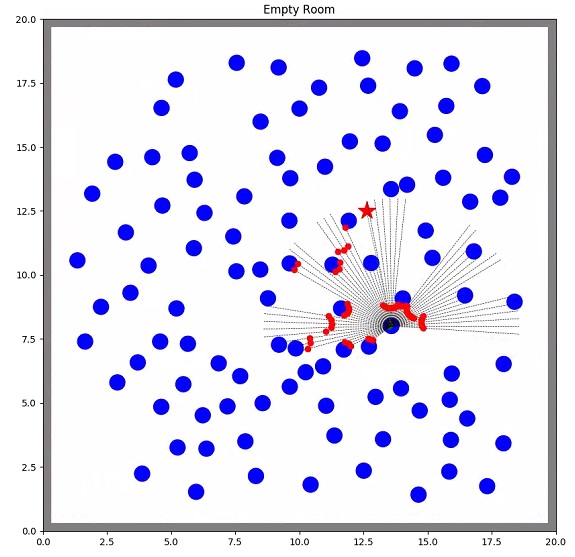}}
    \caption{{\bf Floor maps.} These are some of the floor maps with their sizes in meters. Building 1 and Building 3 maps used for this work are the ones used in \cite{oscar_anais_sensors2021}.}
    \label{floormaps}
\end{figure*}

The training and evaluation in simulation are executed in maps (refer to Fig. \ref{floormaps}), which are the same floor maps used in \cite{oscar_anais_sensors2021}. A new empty scenario has been added to evaluate the performance in presence of only moving obstacles. In this evaluation, the OP-DDPG model is the DDPG trained model without adding the velocity changes or anticipative strategies in the evaluation phase. 

\subsection{Metrics}
To evaluate the models, 3 metrics have been used:

\begin{itemize}
\item $\textbf {Success Rate (SR)}$: Percent of episodes in which the robot can achieve the goal without collision. This metric shows the overall performance.

\item $\textbf {Collision Rate (CR)}$: Percent of episodes that ends in collision. 

\item $ \textbf {Timeout Rate (TR)} $: Percent of episodes where episode time limit is reached. 

\end{itemize}

Other metrics, as for example, the number of steps and the goal distance, give a less global description of the performance.

\subsection{Simulation Results}
Several versions have been evaluated in different conditions to explore the performance and the limits of the anticipative models. In this evaluation, the trained OP-DDPG model is the same for all the simulations. The differences are in the evaluation phase, where the hybrid model and the different anticipative strategies are used. It is important to remark that the AT, ARP and APP models are always used in combination with the hybrid model, rather than the OP-DDPG model. All the evaluations have been performed during 100 episodes as in \cite{chiang2019learning}.

Two ways for evaluation have been considered: using a Euclidean Distance (ED) between the goal and the initial position of the robot in a distance between $5$ and $10 \ m$; and using the approximate path distance (PD) between $5$ and $10 \ m$ of a RRT planner between the initial position and a goal. 

The path distance offers episodes without very large, complex navigation for which the P2P task has not been designed, because the trained model does not use the complete map knowledge. 

The results shown in Table \ref{static} demonstrate that the hybrid model increases the success rate in all the evaluation maps. The AT model increases the performance even more avoiding most of the collisions, but with some timeouts caused by the large-scale local minima. The ARP and APP models are not included because there are not moving obstacles.

\begin{table}[t]
\caption{Model comparison in static environments without moving obstacles for the Euclidean distance (ED) and Path distance (PD) cases.}
\label{static}
\begin{center}
\begin{tabular}{c c c c c c c c c c}
\hline 
\multirow{2}{*}{Environment} & \multicolumn{3}{ c }{OP-DDPG} & \multicolumn{3}{ c }{Hybrid Model} & \multicolumn{3}{ c }{AT} \\ \cline{2-10}
 & \multicolumn{3}{ c }{ SR-CR-TR } & \multicolumn{3}{ c }{ SR-CR-TR } & \multicolumn{3}{ c }{ SR-CR-TR } \\
\hline 
{\bf Building 1 ED} & & 60 - 34 - 6 & & &66 - 11 -23 & & &78 - 0 - 22 \\

{\bf Building 2 ED} &  &50 - 46 - 4&  & &74 - 15 - 11 & & &87 - 1 - 12\\

{\bf Building 3 ED} & &65 - 35 - 0 & & &81 - 11 - 8& & &92 - 0 - 8\\
\hline 
{\bf Building 1 PD} & &75 - 11 - 14& & &88 - 3 - 9& & &94 - 0 - 6\\

{\bf Building 2 PD} & &87 - 13 - 0& & &96 - 1 - 3& & &97 - 0 - 3\\

{\bf Building 3 PD} & &87 - 12 - 1& & &97 - 1 - 2& & &98 - 0 - 2\\
\hline
\end{tabular}
\end{center}
\end{table}

The OP-DDPG model results are worse than the results in \cite{chiang2019learning}, due to the AutoRL optimal hyperparameters \cite{chiang2019learning} are suboptimal in a different implementation. In this work, the results are focused in the advantages of applying the hybrid model and the anticipative strategies, regardless the AutoRL optimization.

Table \ref{dynamic} shows that the anticipative models are robust against the number of agents. Only in environments with a very high density of moving obstacles, like Building 2 with 100 obstacles, the success rate is significantly reduced. The AT+APP model gives the best results for all the combination of models when there are static and moving obstacles.

\begin{table*}[t]
\caption{Model comparison in dynamic environments with 20 and 100 moving agents using the path distance to sample the goals.}
\label{dynamic}
\begin{center}
\begin{tabular}{ccccccccccccccccc}
\hline 
\multirow{2}{*}{Environment} & \multicolumn{3}{ c }{OP-DDPG} & \multicolumn{3}{ c }{Hybrid Model} & \multicolumn{3}{ c }{ARP} & \multicolumn{3}{ c }{APP} & \multicolumn{4}{ c }{APP+AT} \\ \cline{2-17}
 & \multicolumn{3}{ c }{SR-CR-TR} & \multicolumn{3}{ c }{SR-CR-TR} & \multicolumn{3}{ c }{SR-CR-TR} & \multicolumn{3}{ c }{SR-CR-TR} & \multicolumn{4}{ c }{SR-CR-TR}\\
\hline 
{\bf Building 1-20} & &74 - 11 - 15& & &84 - 4 - 12 & & &84 - 6 - 10 & & &85 - 5 - 10& & & &{\bf 96 - 1 - 3}\\

{\bf Building 2-20} & &80 - 20 - 0 & & &89 - 3 - 8& & &88 - 4 - 8& & &89 - 2 - 9& & & &{\bf 93 - 4 - 3}\\

{\bf Building 3-20} & &84 - 16 - 0 & & &88 - 2 - 10& & &88 - 3 - 9 & & &87 - 3 - 11& & & &{\bf 95 - 1 - 4}\\
\hline 
{\bf Building 1-100} & &67 - 16 - 17 & & &84 - 3 - 13 & & &84 - 5 - 11& & &86 - 4 - 10& & & &{\bf 97 - 1 - 2}\\

{\bf Building 2-100} & &61 - 39 - 0 & & &76 - 16 - 8 & & &75 - 16 - 9& & &76 - 14 - 10& & & &{\bf 81 - 17 - 2}\\

{\bf Building 3-100} & &83 - 17 - 0 & & &92 - 0 - 8& & &93 - 1 - 6& & &93 - 1 - 6& & & &{\bf 94 - 3 - 3} \\
\hline
\end{tabular}
\end{center}
\end{table*}

The empty map scenario with only moving obstacles is very useful to evaluate the anticipative strategies. In this way, in Table \ref{empty}, we demonstrate the positive effect of the anticipative strategies in collision avoidance with moving obstacles. In very crowded scenes, where the Euclidean distance for the goal is between $5 \ m$ and $10 \ m$, the success rate improves a 10 percent. 

\begin{table}[h]
\caption{Model comparison in an empty map for 35, 50 and 70 moving agents.}
\label{empty}
\begin{center}
\begin{tabular}{cccccccccc}
\hline 
\multirow{2}{*}{Environment} & \multicolumn{3}{ c }{Hybrid Model} & \multicolumn{3}{ c }{ARP} & \multicolumn{3}{ c }{APP} \\ \cline{2-10}
 & \multicolumn{3}{ c }{SR-CR-TR} & \multicolumn{3}{ c }{SR-CR-TR} & \multicolumn{3}{ c }{SR-CR-TR} \\
\hline 
{\bf E. Map ED-35} & &95 - 5 -0 & & &95 - 5 - 0 & & &{\bf 98 - 2 - 0}&\\

{\bf E. Map ED-50} & &89 - 11 - 0& & &90 - 10 - 0 & & &{\bf 96 - 4 - 0} &\\

{\bf E. Map ED-70} & &81 - 19 - 0 & & &86 - 14 - 0 & & &{\bf 93 - 6 - 1}& \\
\hline
\end{tabular}
\end{center}
\end{table}

The robustness to noise has been evaluated in building 2 with 20 moving agents. Like in AutoRL work, Gaussian noise is considered, $\mathcal{N}(0,\sigma_{lidar})$, for 4 different standard deviations, $\sigma_{lidar}=[0.1, 0.2, 0.4, 0.8]$. The success rate in these cases decreases less than a $3 \%$. 
In these evaluations, the maximum velocity for moving agents is $1.2 \ m/s$ and the SFM is used to simulate a pedestrian as in \cite{chiang2019learning}. 

The last evaluation is in the empty map scenario through intersections between the robot and moving agents for different agent maximum velocities. In this challenging scene, the intersections are forced to check how the robot reacts to avoid collisions. The agents do not avoid obstacles because they use the CVM. 

\begin{figure}[t]
    \centering
    \includegraphics[width=0.70\linewidth]{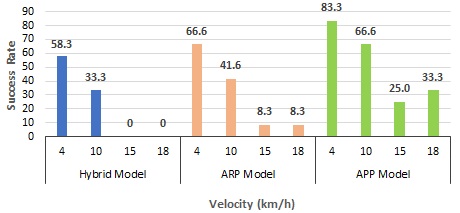}
    \caption{{\bf Success rate variation with agent velocities in the empty map.} The least rate decay occurs in the APP model.}
\vspace{0mm}
\label{velocities}
\end{figure}

This evaluation has been performed using 12 cases with up to 3 agent intersections in different directions regarding robot orientation. These directions are $\theta=[-60^{o},-30^{o},0^{o},30^{o},60^{o}]$. The results in Fig. \ref{velocities} show less decay for the APP model. For more than $18 \ km/h$ it is almost impossible to avoid collisions because in the Lidar's range, the robot has less than a second to react.
\subsection{Real-Life Experiments}

Real indoor experiments have been performed using the Helena robot. Helena is a transporter robot with a Pioneer P3-DX base and a RS-LiDAR-16. The Lidar observations have been adapted to be entries of the trained policy. 

Two experiments have been performed to check the feasibility of the APP model with people in real environments:
\begin{itemize}
\item $\textbf {Frontal Encounter (FE)}$: A person and the robot cross paths walking in opposite directions.
\item $\textbf {Perpendicular Encounter (PE)}$: A person and the robot cross paths walking in perpendicular directions.
\end{itemize}

\begin{figure}[t!]
    \centering
    \subfigure[Hybrid model t]{\includegraphics[height=0.16\textwidth]{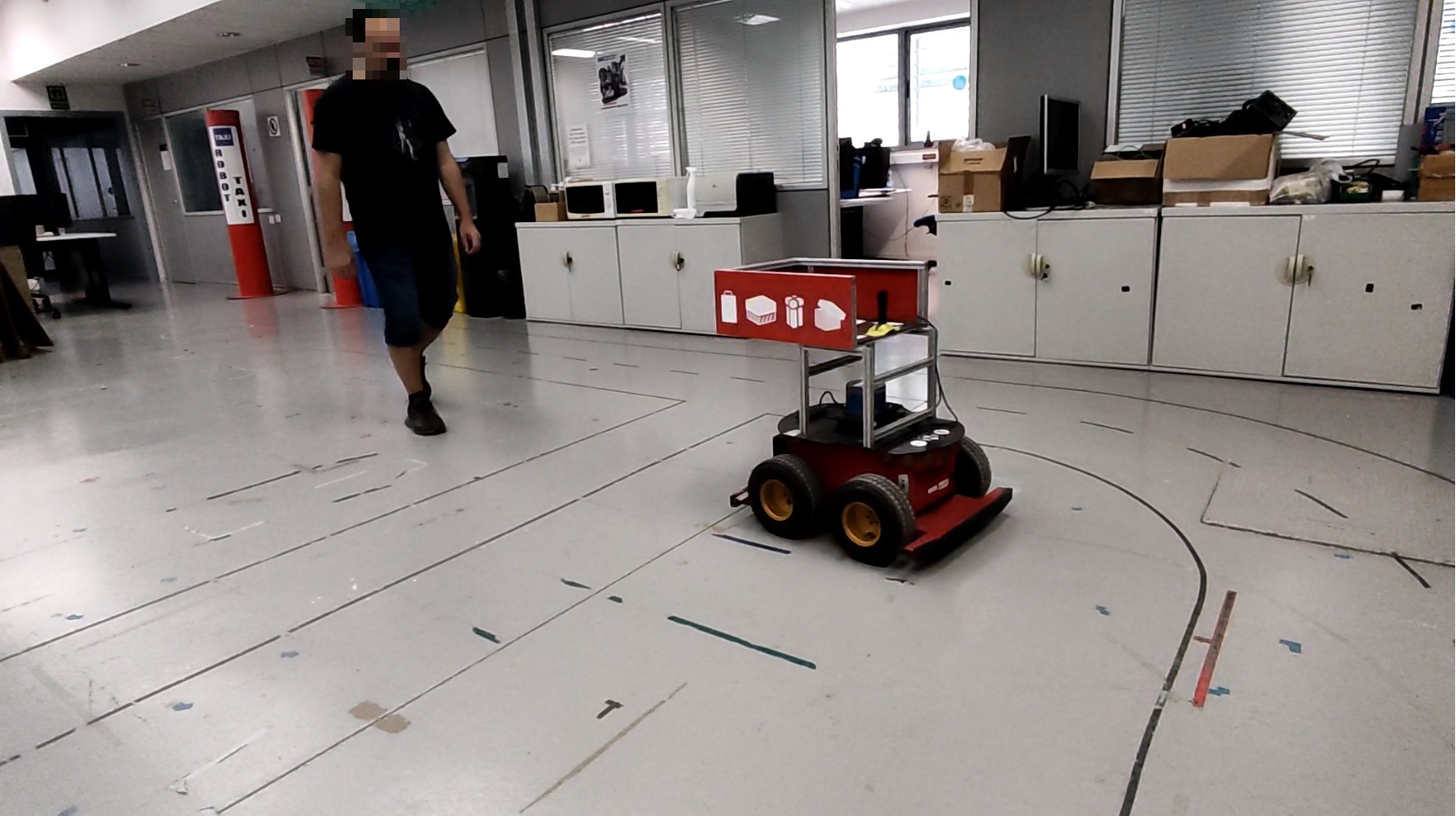}}
    \subfigure[Hybrid model t+1]{\includegraphics[height=0.16\textwidth]{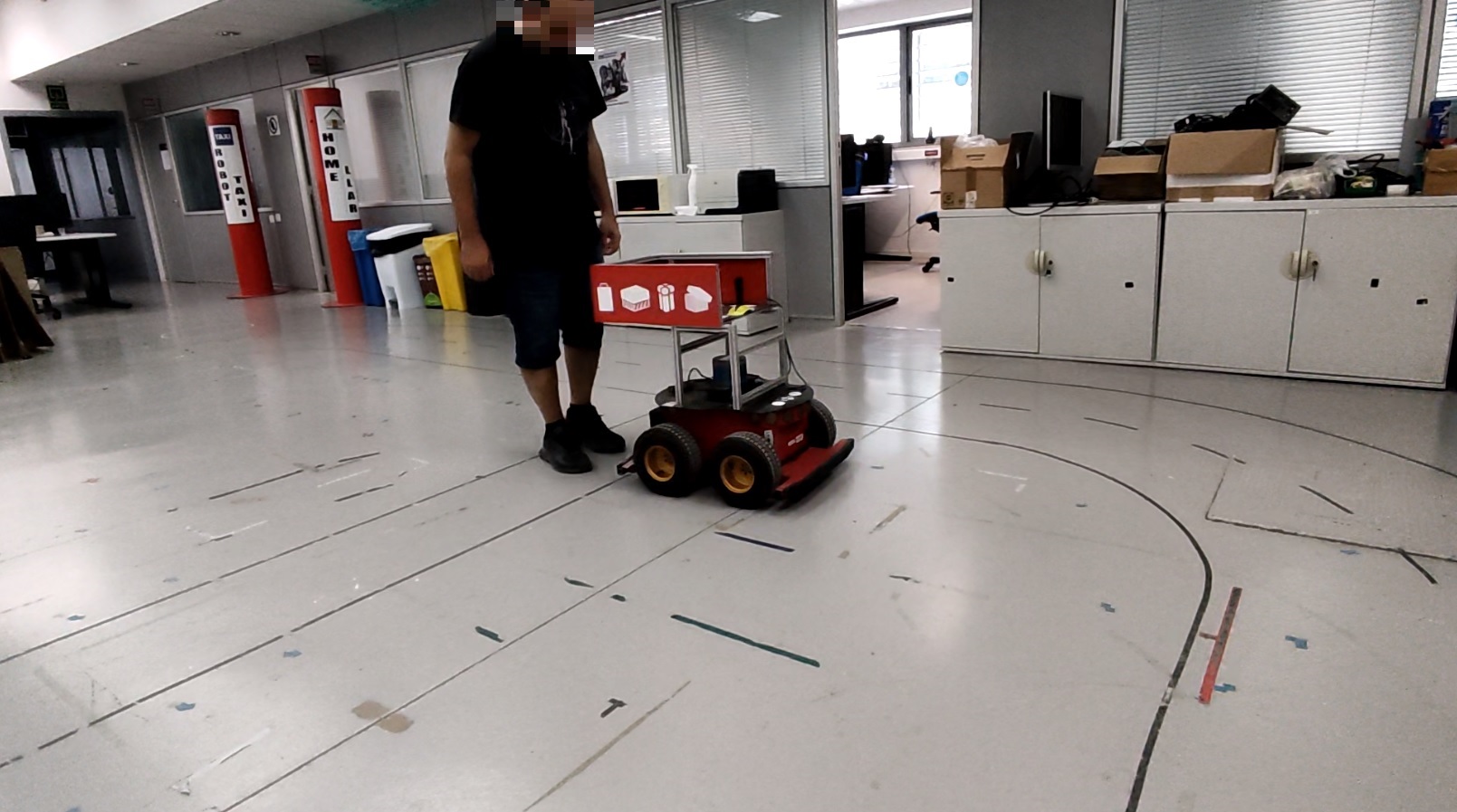}}
    \subfigure[Hybrid model t+2]{\includegraphics[height=0.16\textwidth]{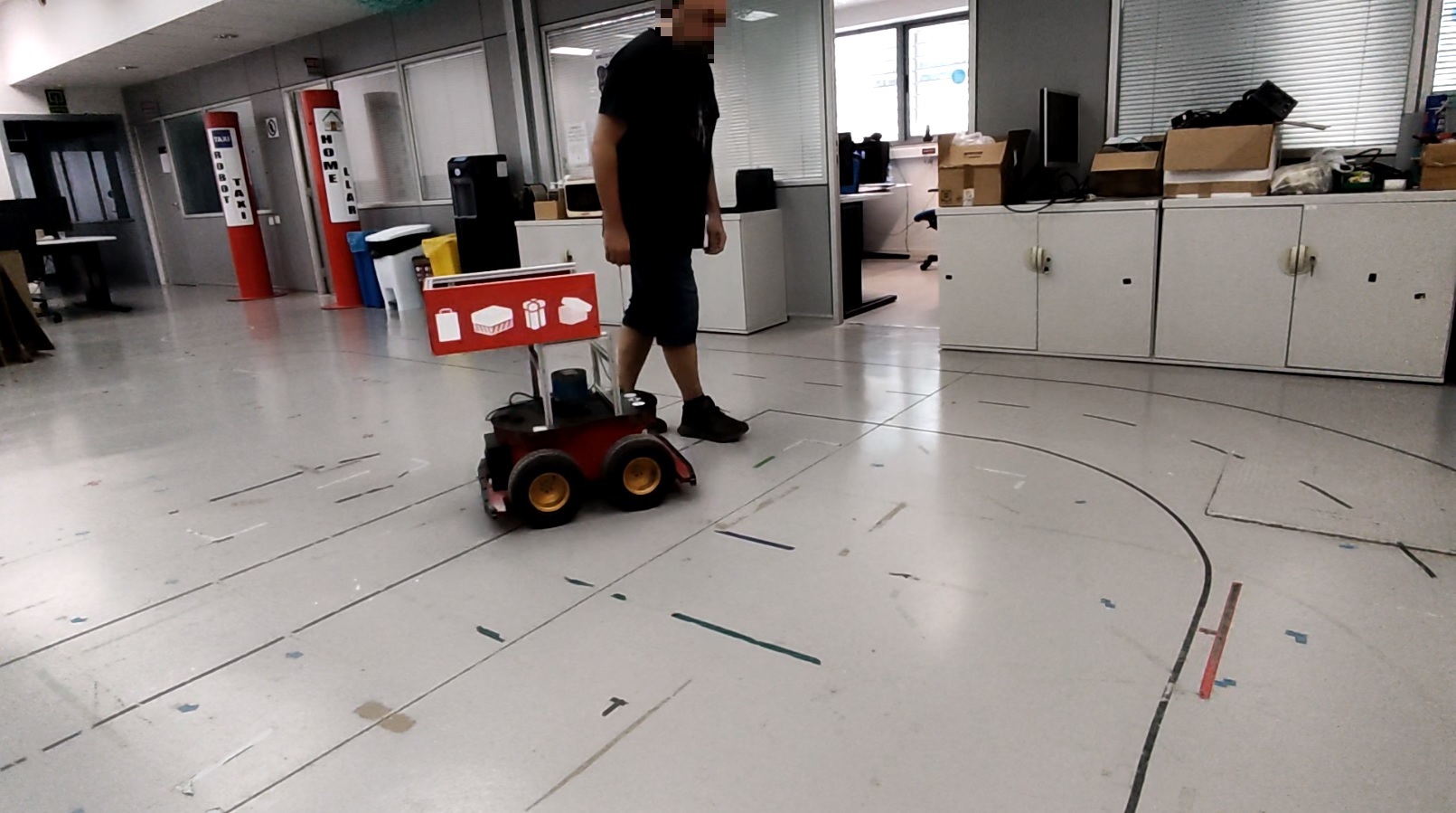}}
    \subfigure[APP model t]{\includegraphics[height=0.16\textwidth]{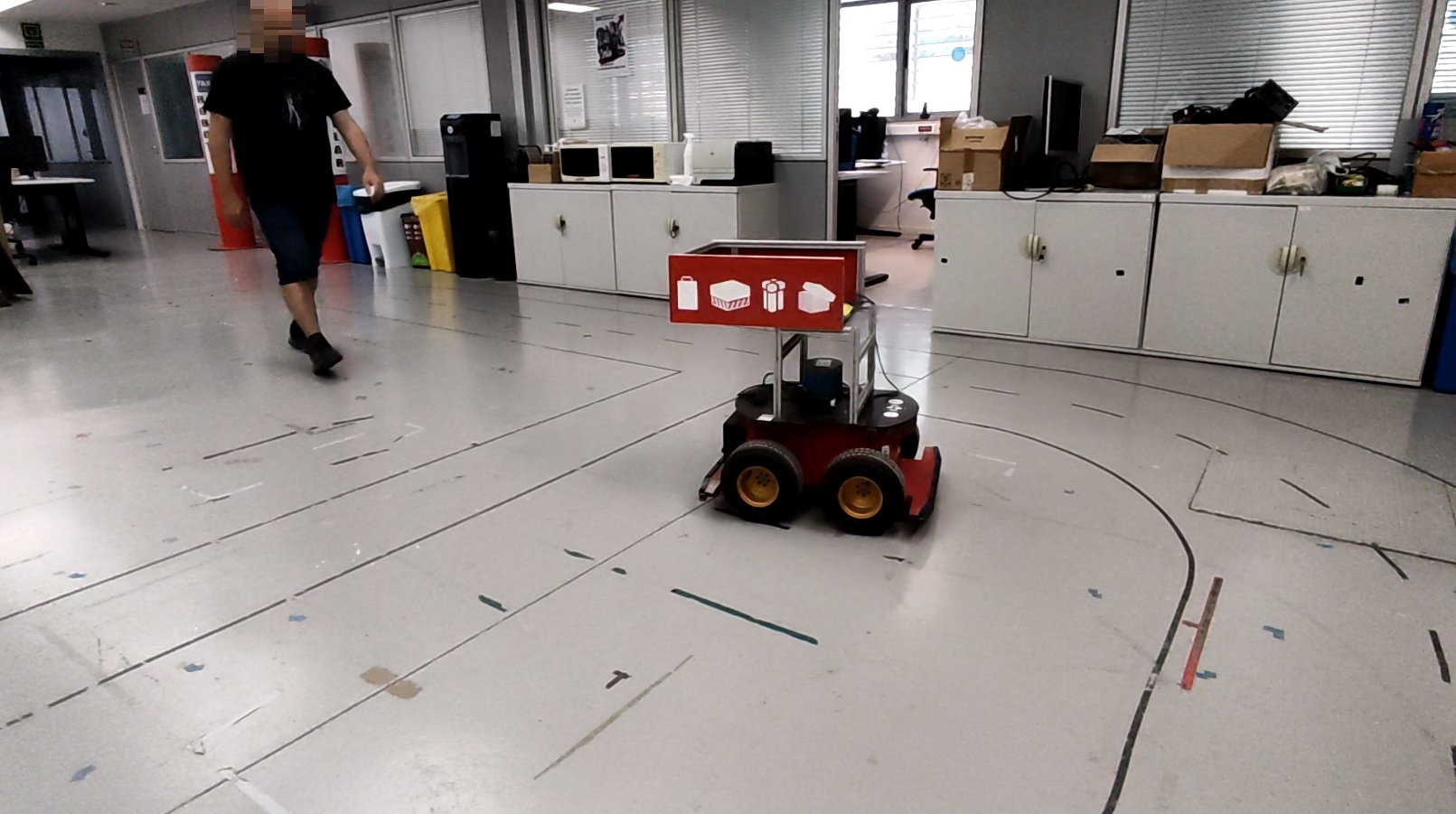}}
    \subfigure[APP model t+1]{\includegraphics[height=0.16\textwidth]{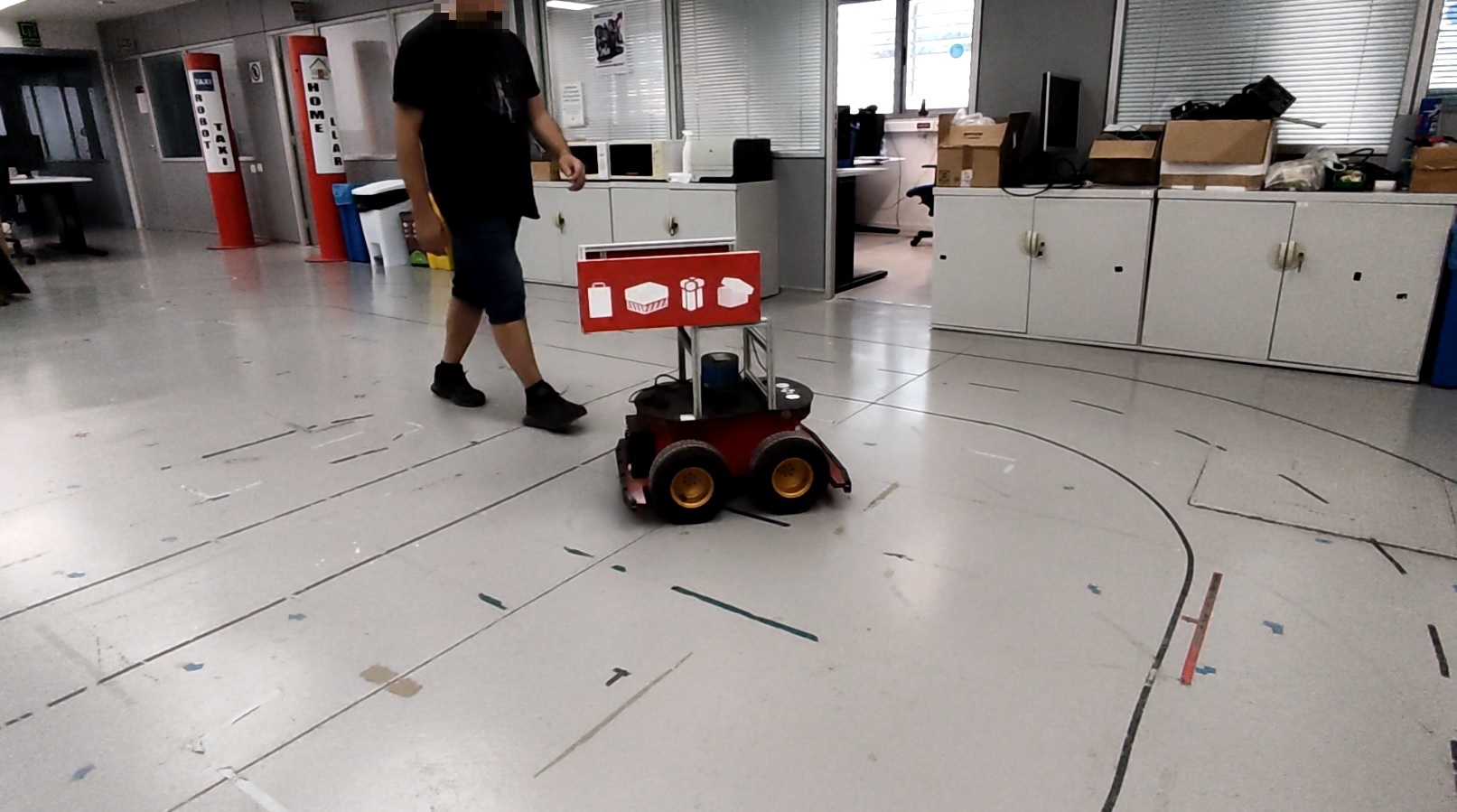}}
    \subfigure[APP model t+2]{\includegraphics[height=0.16\textwidth]{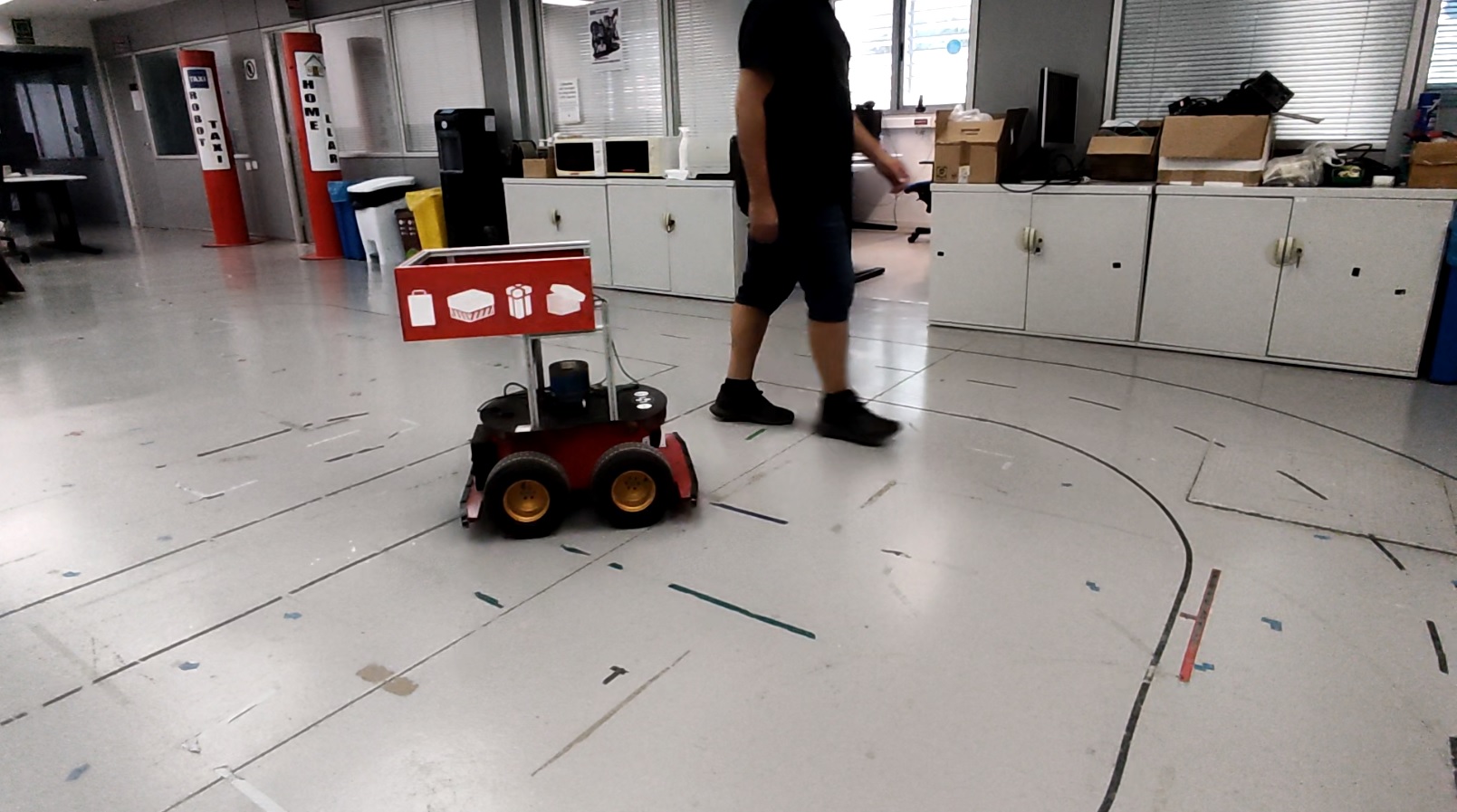}}
    \caption{{\bf FE experiment.} From time t to t+2 the images show the Helena robot behavior when the robot uses the hybrid model and the APP model.} 
\vspace{0mm}
\label{exp_person}
\end{figure}

\begin{figure}[t!]
    \centering
    \subfigure[Hybrid model t]{\includegraphics[height=0.16\textwidth]{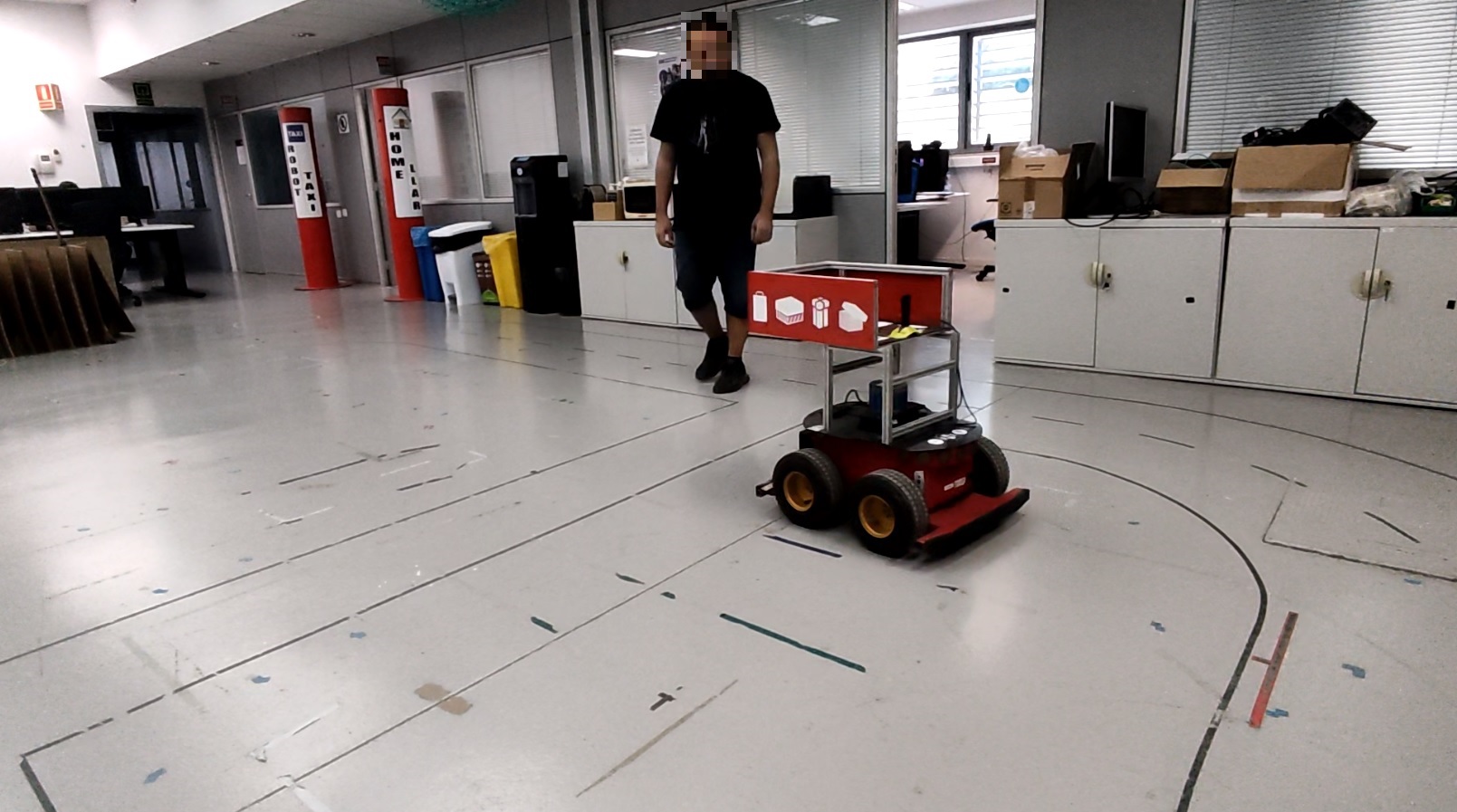}}
    \subfigure[Hybrid model t+1]{\includegraphics[height=0.16\textwidth]{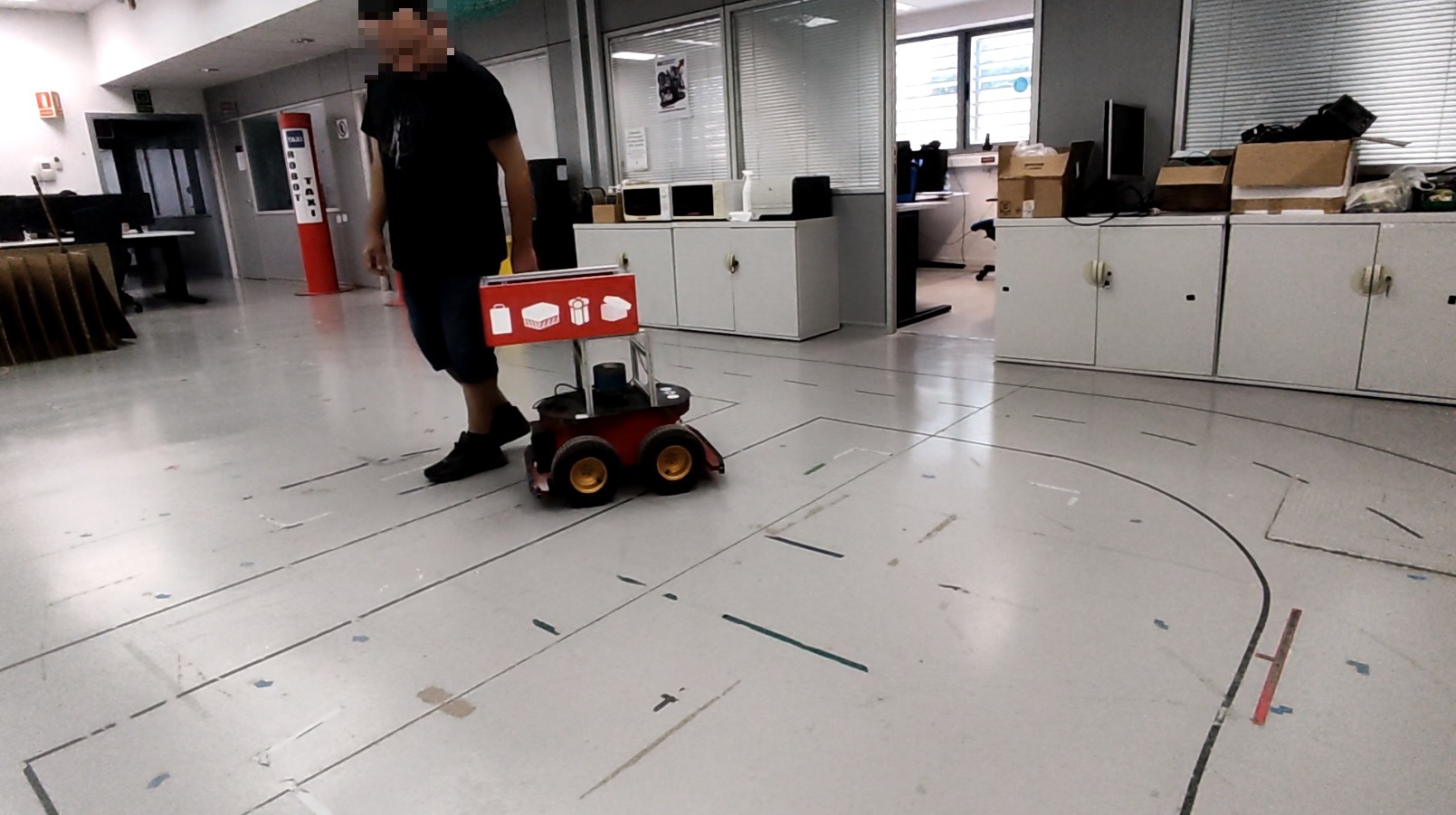}}
    \subfigure[Hybrid model t+2]{\includegraphics[height=0.16\textwidth]{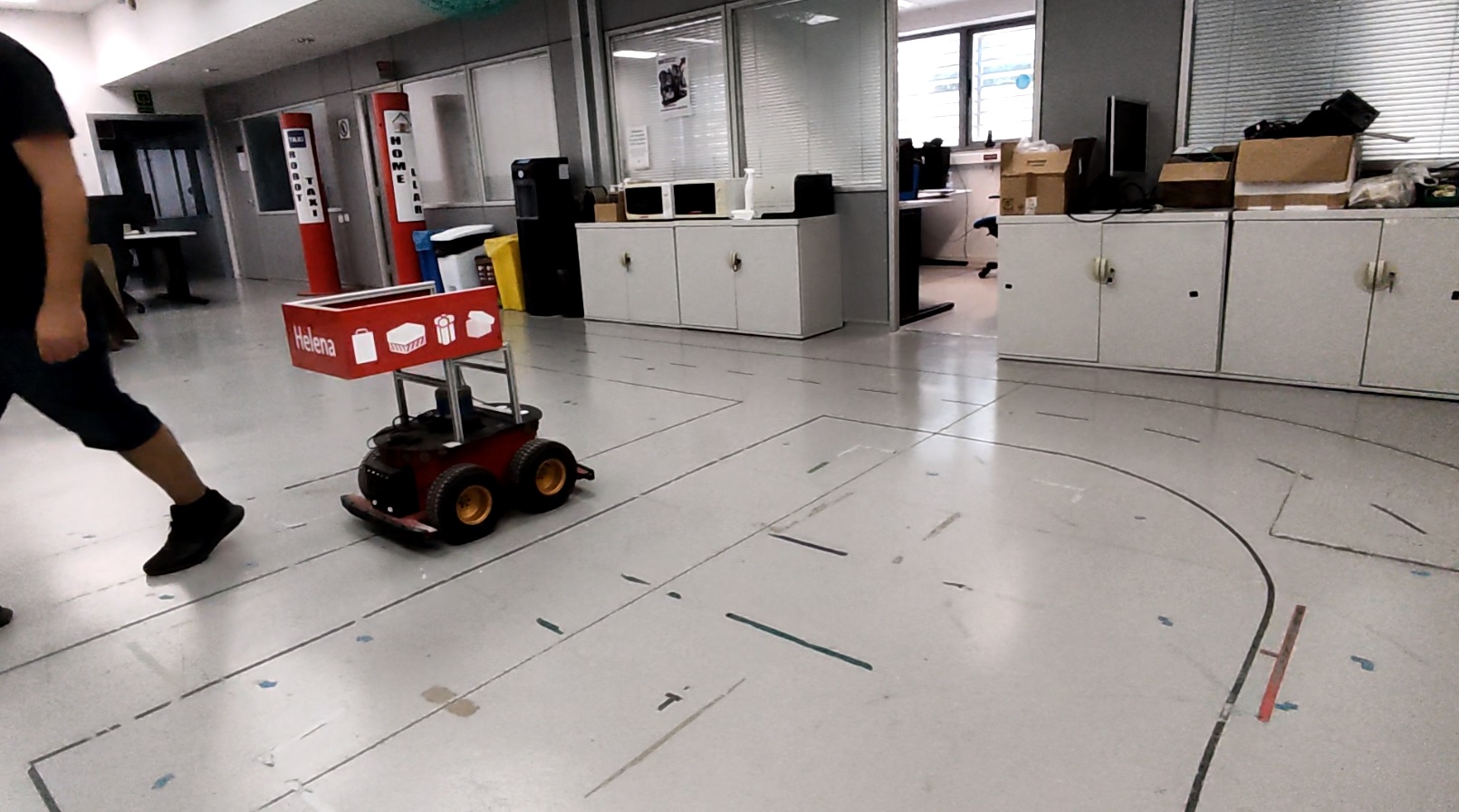}}
    \subfigure[APP model t]{\includegraphics[height=0.16\textwidth]{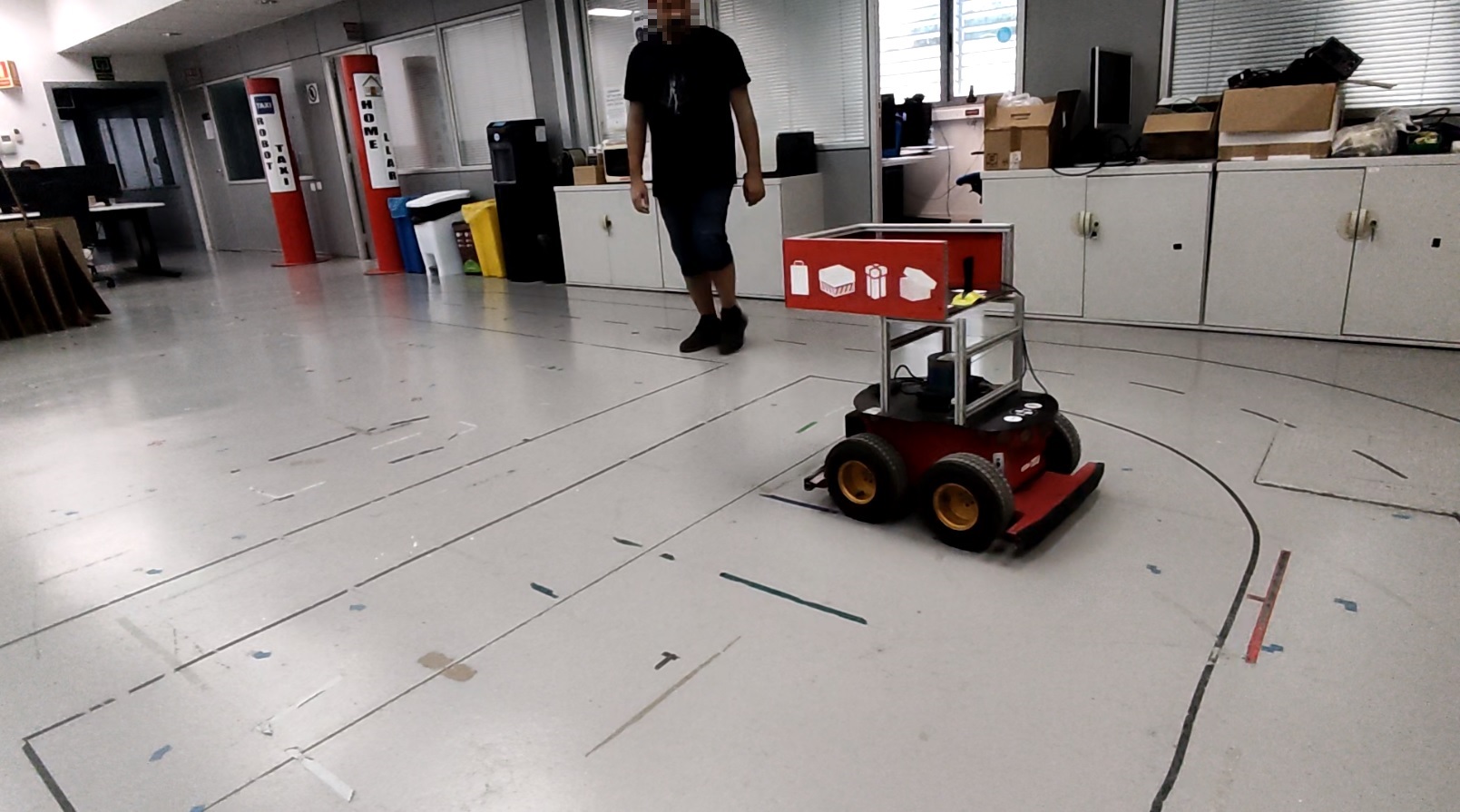}}
    \subfigure[APP model t+1]{\includegraphics[height=0.16\textwidth]{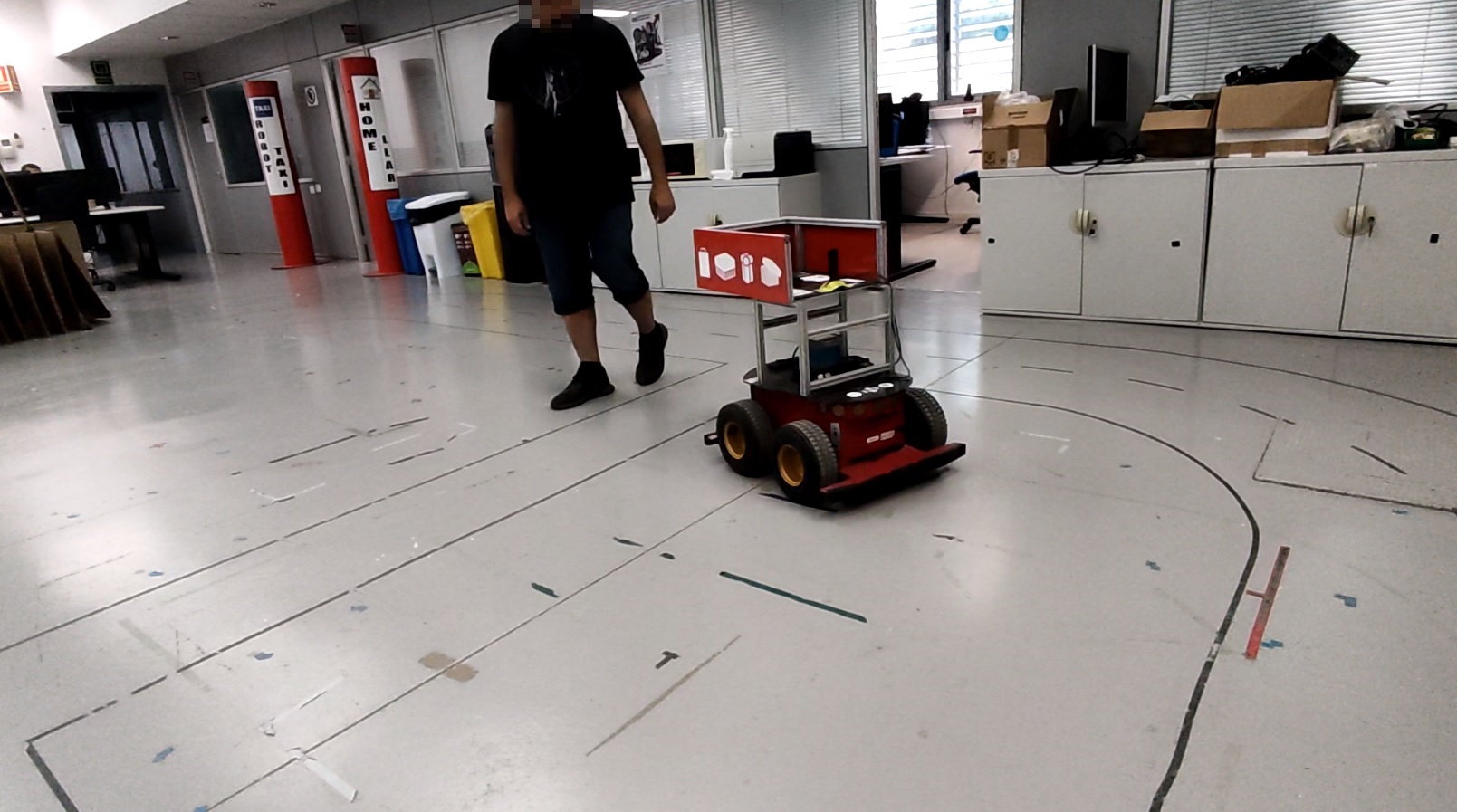}}
    \subfigure[APP model t+2]{\includegraphics[height=0.16\textwidth]{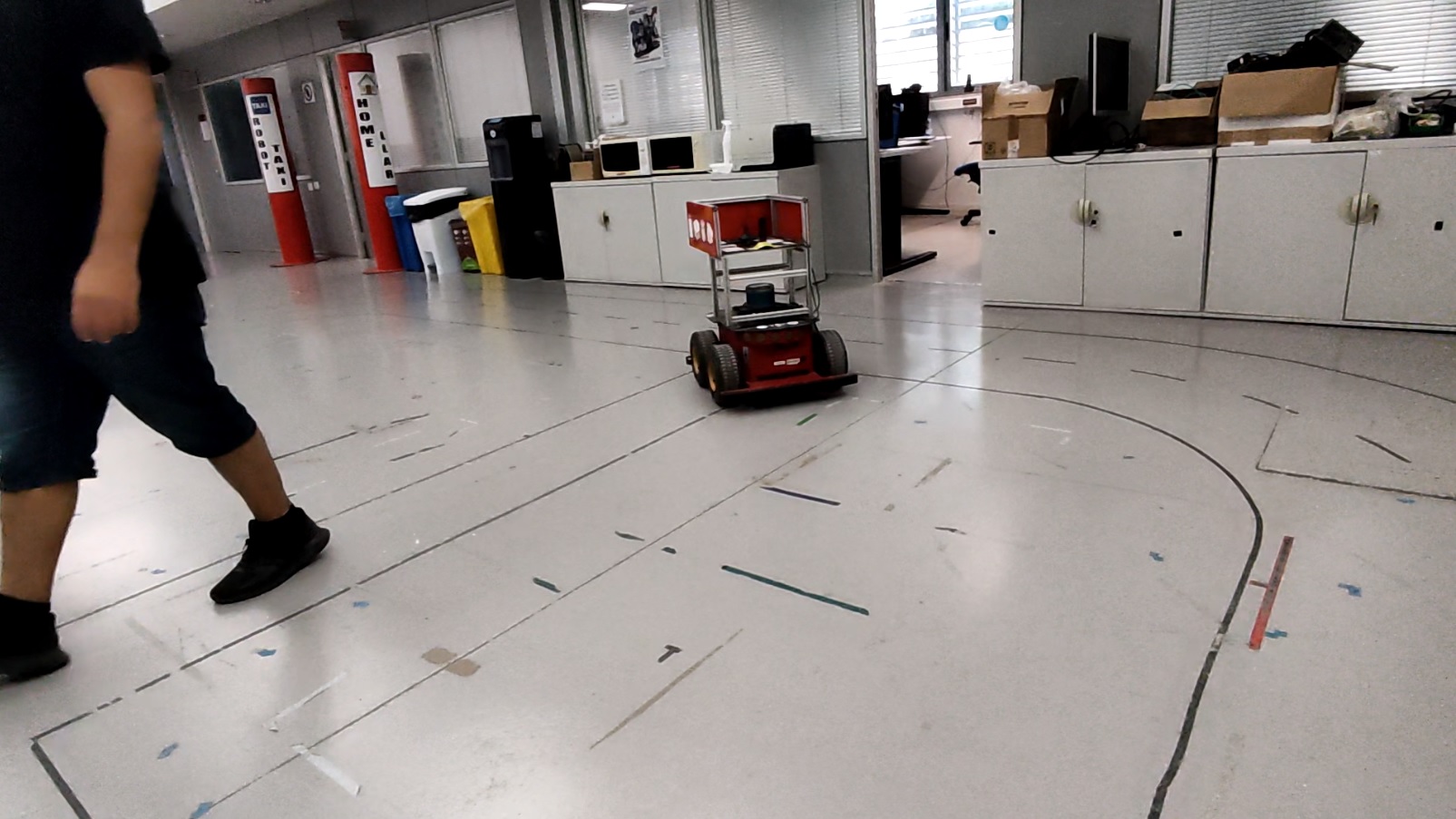}}
    \caption{{\bf PE experiment.} From time t to t+2 the images show the Helena robot behavior when the robot uses the hybrid model and the APP model.} 
\vspace{0mm}
\label{exp_person_2}
\end{figure}

\begin{figure}[t!]
    \centering
    \subfigure[NS-DWA]{\includegraphics[height=0.17\textwidth]{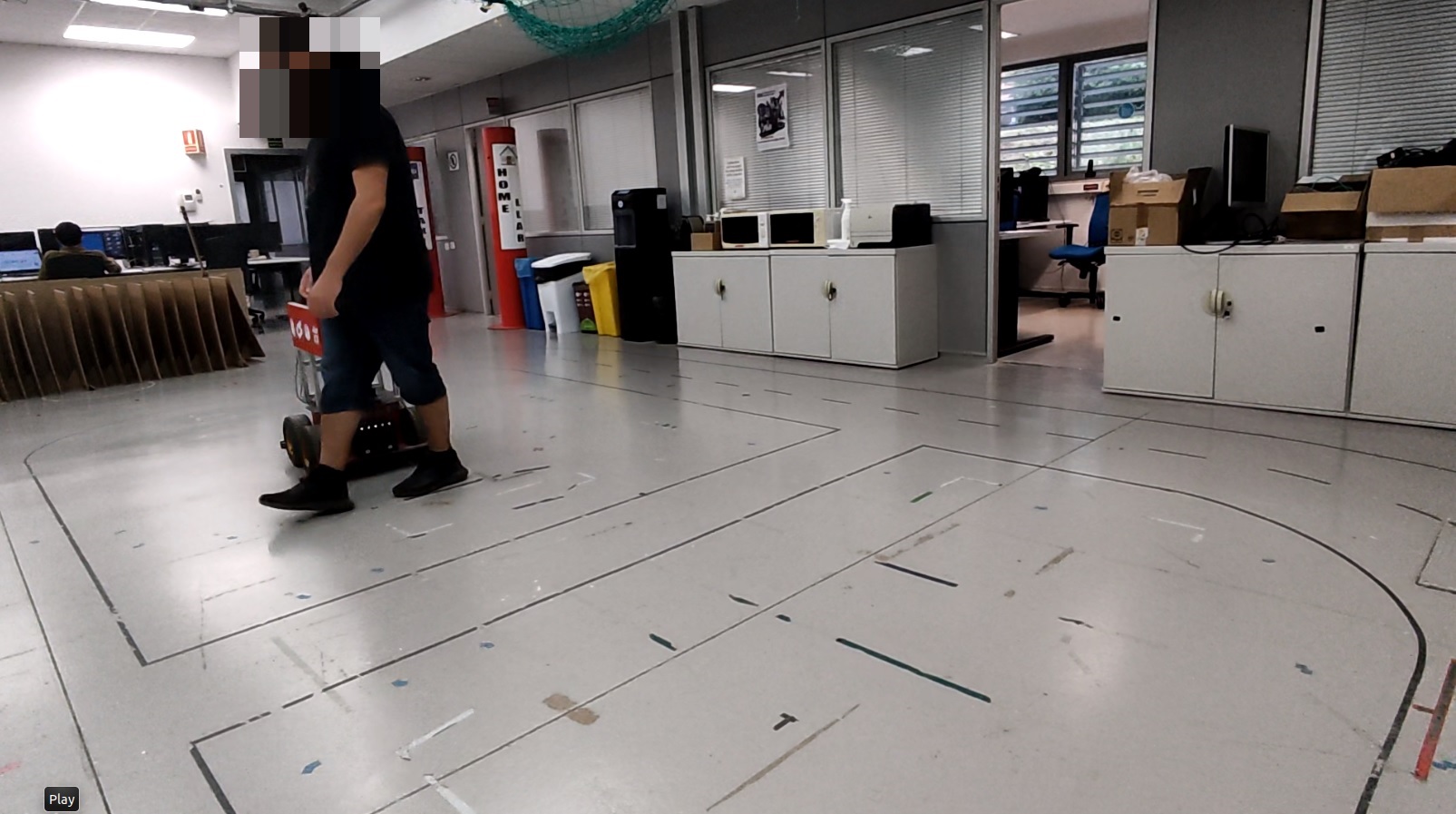}}
    \subfigure[NS-DWA + APP]{\includegraphics[height=0.17\textwidth]{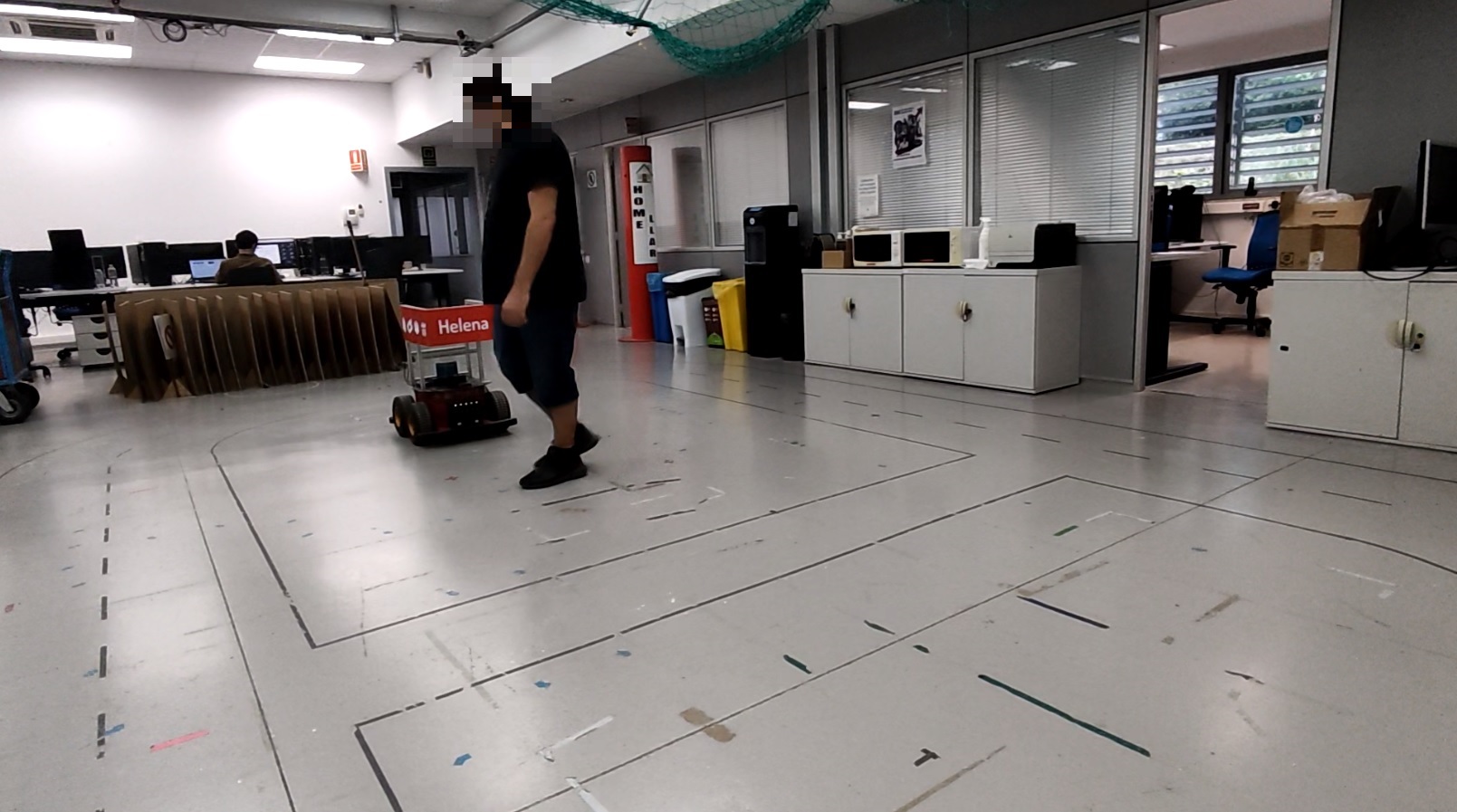}}
    \caption{{\bf PE experiment with NS-DWA.} The APP model stops the robot before the encounter.} 
\vspace{0mm}
\label{exp_person_3}
\end{figure}

In Fig. \ref{exp_person}, the FE experiments are shown. With the hybrid model the robot only reacts close to the person. Using the APP model, the robot reacts around $2 \ m$ before the person arrives. In Fig. \ref{exp_person_2}, the PE experiments show that the robot using the APP model not only reacts before, but chooses to navigate behind the person to avoid collisions which is a more efficient solution.

In Fig. \ref{exp_person_3}, the PE experiment is performed using the ROS Navigation Stack with DWA (NS-DWA) and the APP model through anticipative circles created in the costmap. The NS-DWA reduces the robot velocity too close of the person but using the APP model the robot stops before reach the person. Overall, the hybrid model robot movements are more natural than the ones observed with the NS-DWA when a person is close to the robot, 
because the robot does not stop to replan its path, avoiding in a better way the frozen robot problem. The minimum distance to the robot increases and the navigation time is reduced (refer to Table \ref{time_navigation}).

\begin{table}[h!]
\caption{
Navigation time and minimum distance to the person for the PE experiment. Helena maximum velocity is $0.5 \ m/s$ and the goal distance is 6.5 $m$.}
\label{time_navigation}
\begin{center}
\begin{tabular}{ccccc}
\hline 
& {\bf NS-DWA} & {\bf NS-DWA+APP} & {\bf Hybrid Model} & {\bf APP model}\\
\hline
{\bf Navigation Time (s)} & 18.8 & 19.6 & 20.0 & {\bf 17.4}\\

{\bf M. distance (m)} & 0.35 & {\bf 0.57} & 0.36 & 0.53\\
\hline
\end{tabular}
\end{center}
\end{table}
\subsection{Implementation Details}
In this subsection, the main features used in our implementation are detailed. All the hyperparameters used in OP-DDPG have been taken from \cite{francis2020long}.

For the neural network, the critic joint, critic observation and actor layer widths are $(607,242)\times(84)\times(241,12,20)$ with ReLU as activation function. For the reward, the parameters $\boldsymbol{\theta}_{r_{P2P}}^{T}$ are $(-0.43, 0.38, -57.90, 0.415, 0.67, 62.0)$. At each step, the number of observations considered is $\theta_{n}=1$. The noise parameters are the ones described in \cite{francis2020long}. All the results have been obtained with the same trained model during 2 million steps and 500 steps as maximum episode length. 

The implementation has been performed in PyTorch using the Adam optimizer, the Gym library and the Shapely library. 

The force parameters used in SFM are $A_{i}=0.7$ and $B_{i}=10/17$. For the ARP and APP models the propagation number of steps is $n=20$. The anticipative circles in APP model are created every 5 steps with $\Sigma=0.01$ $m$. This value produces circles with a shorter diameter than corridors where the robot and a moving agent can navigate. 

For all simulations, each simulation step is $0.2 \ s$. The agents and the robot are circles with a radius of 0.3 $m$. To perform real experiments, the implementation has been transferred to the Robot Operation System (ROS) middleware and the people velocities are obtained through the Spencer tracker \cite{linder2016multi}. 
\section{Conclusions}
In this work, we have analyzed different methods to avoid collisions with people in a reactive and anticipative way, using a model trained in a static environment. 

In scenarios with static obstacles, the hybrid model obtains better results than the OP-DDPG model. In scenarios with static and moving obstacles, the combination between the hybrid model and 3 different anticipative strategies to prevent collisions in a $5 \ m$ range, obtains better results than the hybrid model. 


The AT model results show an important improvement avoiding collisions with static obstacles. For dynamic environments, the ARP and APP models improve the performance of the other models, because the robot anticipates the collision and reacts before the collision happens. Specifically, the APP model can avoid in some situations high-velocity agents with speeds of up to $18 \ km/h$. Furthermore, the APP improves the robot anticipation in real experiments with people for the hybrid model and the NS-DWA. These anticipative models could be applied for velocities greater than $18 \ km/h$ with a longer Lidar's range.


%
%

\bibliographystyle{splncs03}
\bibliography{robot_2022.bib}

\end{document}